\documentclass[letterpaper, 10 pt, conference]{ieeeconf}

\IEEEoverridecommandlockouts
\usepackage[T1]{fontenc}
\usepackage[utf8]{inputenc}
\usepackage[english]{babel}
\usepackage{graphicx}
\usepackage{url}
\usepackage{cite}
\usepackage{amsmath}
\usepackage{amssymb}
\usepackage{booktabs}
\usepackage{balance}
\usepackage{microtype}
\usepackage{multirow}
\usepackage{adjustbox}
\usepackage[hidelinks,breaklinks=true]{hyperref}

\title{\LARGE \bf
The Impact of Detector Architecture, Model Scale, and Scene Density on Multi-Pedestrian Tracking
}

\author{Pedro Manoel Herminio Alves$^{1,*}$, Dalton Cézane Gomes Valadares$^{1,2}$, Cláudio de Souza Baptista$^{1}$%
\thanks{$^{1}$Federal University of Campina Grande, Brazil}%
\thanks{$^{2}$Federal University of Paraíba, Brazil}%
\thanks{$^{*}$pedro.alves@copin.ufcg.edu.br}%
}

\begin{document}

\maketitle
\thispagestyle{empty}
\pagestyle{empty}

\begin{abstract}
Pedestrian tracking in video is important for applications such as surveillance and scene analysis, but it becomes more challenging in crowded scenarios with frequent occlusions and rapid motion. This paper analyzes how detector architecture, model scale, scene density, and inference regime affect the tracking of multiple pedestrians. We compare the YOLO11, RT-DETR, and YOLO26 detector families with the ByteTrack and BoT-SORT trackers on sequences from the MOT17, MOT20, and SOMPT22 benchmarks. For YOLO26, we also compare traditional overlap filtering via Non-Maximum Suppression (NMS) with end-to-end inference. The RT-DETR family achieved the highest mean scores on the HOTA tracking quality metric across all three benchmarks with both trackers, with a more pronounced advantage on MOT20, albeit at a higher computational cost. On MOT17 and SOMPT22, YOLO26 variants, especially YOLO26-L with NMS, offered favorable combinations of quality and speed compared with RT-DETR. In the YOLO26 inference comparison, the end-to-end mode did not yield consistent tracking gains over the NMS-based regime. Taken together, these results provide a practical guide for selecting detectors and trackers for this task based on application priorities.
\end{abstract}

%=============================================================
\section{INTRODUCTION}
%=============================================================

In many scenarios, automatic video analysis relies on the ability to track objects over time. Among these tasks, multiple object tracking (MOT) plays a key role in applications such as surveillance, autonomous driving, robotics, and the analysis of complex scenes~\cite{luo2021multiple}. In the specific case of pedestrians, the problem is particularly relevant because people interact, occlude one another, frequently enter and exit the frame, and may have similar appearances. In these situations, the challenge is not only to detect individuals in isolated frames but to correctly preserve each person’s identity throughout the entire video sequence~\cite{dendorfer2020mot20,simsek2022sompt22}.

The most widely used formulation for this problem is tracking-by-detection. In this approach, an object detector processes each frame and produces bounding boxes with confidence scores. Next, a tracker receives these detections and determines which bounding boxes across consecutive frames correspond to the same person, thereby forming coherent temporal trajectories~\cite{bewley2016sort,zhang2022bytetrack}. This decomposition has become popular due to its simplicity and modularity, but it introduces a direct dependency between the detector’s characteristics and the final tracking performance. In tracking pipelines, the detector architecture, model scale, and type of output produced can alter the set of boxes delivered to the tracker, affecting coverage, confidence distribution, and temporal stability.

Another decisive factor is scene density. In less dense environments, temporal association tends to be simpler, as there are fewer occlusions and less ambiguity among targets. In crowded scenes, on the other hand, overlaps, trajectory interruptions (fragmentations), and the likelihood of identity switching increase~\cite{simsek2022sompt22,aharon2022botsort}. Thus, the same choice of detector and tracker can behave quite differently depending on the visual context, significantly impacting the system’s overall accuracy and stability~\cite{alkandary2025comparative,nguyen2025rtdetrdeepsort}.

In this context, the detector’s inference scheme also warrants attention. In conventional detectors, predictions undergo a post-processing step called Non-Maximum Suppression (NMS), which is responsible for removing redundant boxes. In end-to-end detectors, the model itself directly produces a final set of predictions, reducing or eliminating the need for this external stage~\cite{sun2021e2e,wang2024yolov10,zhao2024rtdetr}. Although this distinction has received increasing attention at the detection level, it is still unclear to what extent it translates into actual gains in tracking.

Motivated by this scenario, we investigated how detector architecture, model scale, scene density, and inference regime affect pedestrian tracking. To this end, we compared the YOLO11, RT-DETR, and YOLO26 detectors with the ByteTrack and BoT-SORT trackers on sequences from the MOT17, MOT20, and SOMPT22 benchmarks. In YOLO26, we also performed an ablation study comparing inference with NMS and end-to-end inference.

In summary, this work makes three main contributions. First, it presents a unified comparison of three recent families of detectors and two trackers widely used in pedestrian MOT. Second, it compares the performance of these combinations across benchmarks with different average pedestrian densities. Third, it evaluates, from a tracking perspective, the effect of inference with NMS and end-to-end inference in YOLO26. Additionally, the code and experimental artifacts are publicly available in the repository\footnote{\url{https://github.com/pedro-manoel/mot-detectors}}.

\section{RELATED WORK}

Recent studies consistently show that tracking performance heavily depends on the detector used. Alkandary et al.~\cite{alkandary2025comparative} compared different versions of the YOLO family, from v3 to v10, combined with the DeepSORT and StrongSORT trackers, in experiments conducted on the MOT17 and MOT20 benchmarks. Similarly, Quan et al.~\cite{quan2024people} analyzed different combinations of YOLO-family detectors and trackers for pedestrian tracking on the same benchmarks. Taken together, these studies show that the relationship between detector and tracker is complex: different combinations can lead to quite distinct results. At the same time, they focus primarily on the choice of detector--tracker pair, without directly examining how the detector’s output format affects tracking and without including a scenario complementary to MOT17 and MOT20.

Another recent line of research has focused on newer detectors and new ways of producing detections. González Hernández et al.~\cite{gonzalez2025analysis} compared variants of the YOLO and RT-DETR families at the detection level, while Nguyen et al.~\cite{nguyen2025rtdetrdeepsort} integrated RT-DETR into a complete tracking pipeline with DeepSORT, showing that Transformer-based detectors can also be successfully used in MOT applications. In parallel, Sun et al.~\cite{sun2021e2e} discuss the methodological importance of eliminating NMS in modern detection, and Zhao et al.~\cite{zhao2024rtdetr} demonstrate the feasibility of real-time Transformer detectors.

Previous studies show that detector choice influences tracking performance and highlight advances in end-to-end detection. However, it remains unclear how these different choices affect pedestrian tracking when analyzed together under a single experimental protocol, considering distinct detector families, different model scales, different methods of generating detections, and benchmarks with contrasting density levels. To the best of our knowledge, this is the first study to systematically evaluate the YOLO11, RT-DETR, and YOLO26 families across their available scales on MOT17, MOT20, and SOMPT22 using the ByteTrack and BoT-SORT trackers.

%=============================================================
\section{METHODOLOGY}
%=============================================================

\subsection{Experimental Design}

\begin{figure*}[t]
    \centering
    \includegraphics[width=1\textwidth]{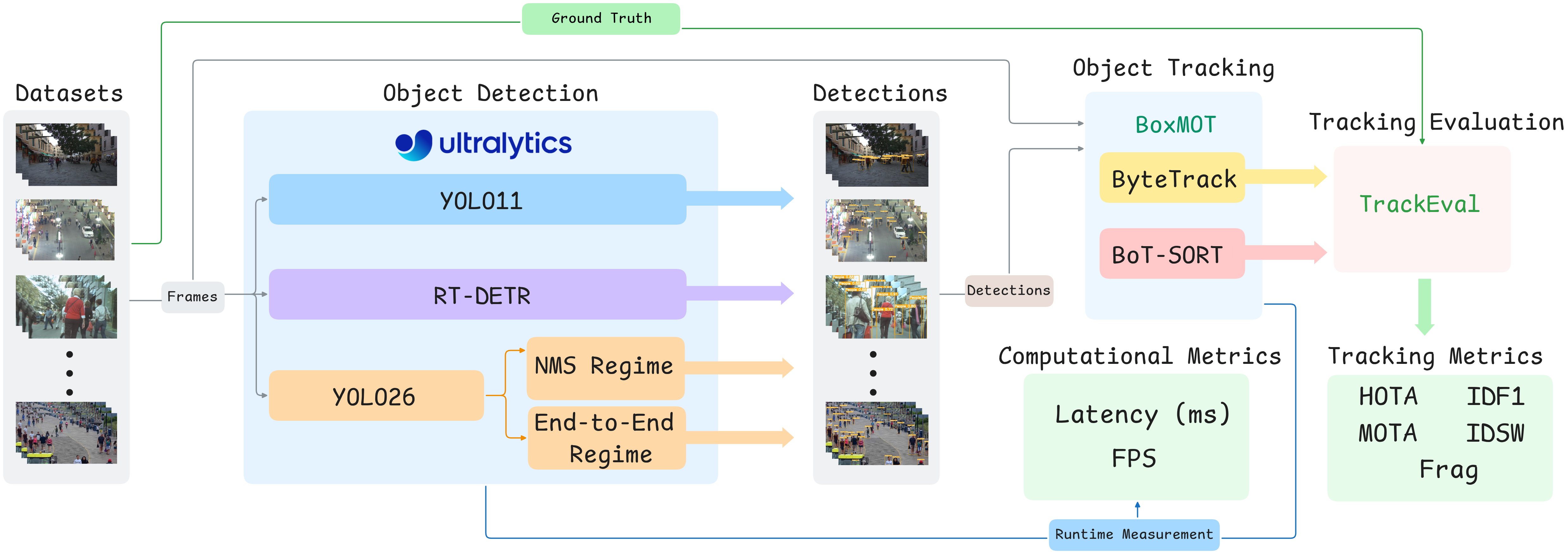}
    \caption{Experimental protocol.}
    \label{fig:pipeline}
\end{figure*}

The experimental protocol was designed to analyze four aspects of pedestrian tracking in \textit{tracking-by-detection} pipelines: \textit{(i)} the effect of the detector family, \textit{(ii)} the effect of the model scale, \textit{(iii)} performance across benchmarks with different average pedestrian densities, and \textit{(iv)} the effect of the detector’s inference regime. To this end, we compared YOLO11, RT-DETR, and YOLO26, combined with the ByteTrack and BoT-SORT trackers. For YOLO26, we also performed an ablation study comparing inference with NMS and end-to-end inference, as illustrated in Figure~\ref{fig:pipeline}.

All models were evaluated using public weights pre-trained on COCO, without fine-tuning on the benchmarks used. This choice keeps the protocol more controlled and closer to a practical deployment scenario, in which the system is evaluated as originally released. Thus, the comparison focuses on the choices of detector, scale, tracker, and inference regime, without additional benchmark-specific training. Since the models were not trained specifically on these tracking datasets, the results demonstrate how well the pre-trained weights perform in new scenarios, which may benefit some detector families more than others.

\subsection{Detectors and Trackers}

The detectors were chosen to represent complementary real-time detection alternatives, and all were run within the Ultralytics framework\footnote{\url{https://github.com/ultralytics/ultralytics}}. All detectors were configured with an input size of 640 pixels and a confidence threshold of 0.25, with detections restricted to the person class. An IoU threshold of 0.7 was used for NMS-based inference. YOLO11 is a convolutional variant of the YOLO family that uses an NMS-based inference pipeline. RT-DETR~\cite{zhao2024rtdetr} is a Transformer-based alternative that uses a hybrid encoder with global attention and a decoder with bipartite matching, formulated end-to-end. YOLO26~\cite{chakrabarty2026yolo26} was included because it allows, within the same convolutional family, a controlled comparison between NMS outputs and end-to-end outputs. Table~\ref{tab:models} summarizes the number of parameters and floating-point operations (FLOPs) for each evaluated model.

\begin{table}[t]
\centering
\caption{Parameters and FLOPs of the evaluated detectors.}
\label{tab:models}
\scriptsize
\setlength{\tabcolsep}{4pt}
\begin{tabular}{lcc}
\toprule
Model & Parameters (Millions) & FLOPs (Billions) \\
\midrule
YOLO11-N & 2.6 & 6.5 \\
YOLO11-S & 9.4 & 21.5 \\
YOLO11-M & 20.1 & 68.0 \\
YOLO11-L & 25.3 & 86.9 \\
YOLO11-X & 56.9 & 194.9 \\
\midrule
RT-DETR-L & 32.1 & 103.8 \\
RT-DETR-X & 65.6 & 222.8 \\
\midrule
YOLO26-N & 2.4 & 5.4 \\
YOLO26-S & 9.5 & 20.7 \\
YOLO26-M & 20.4 & 68.2 \\
YOLO26-L & 24.8 & 86.4 \\
YOLO26-X & 55.7 & 193.9 \\
\bottomrule
\end{tabular}
\end{table}

For each family, we evaluated the officially released scales whenever available (\textit{n}, \textit{s}, \textit{m}, \textit{l}, \textit{x}), which correspond, respectively, to \textit{nano}, \textit{small}, \textit{medium}, \textit{large}, and \textit{extra-large}. Throughout the text and tables, model names use uppercase scale suffixes separated by a hyphen (e.g., YOLO11-L, YOLO26-N, RT-DETR-X). For YOLO26, we add the tags (NMS) and (E2E) to distinguish the NMS-based and end-to-end inference regimes, respectively.

The trackers were chosen to represent two temporal association strategies. ByteTrack~\cite{zhang2022bytetrack} relies primarily on position, motion, and reuse of low-confidence detections, making it a simple and efficient alternative. BoT-SORT~\cite{aharon2022botsort} adds motion compensation and appearance information via OSNet~\cite{zhou2019osnet}-based re-identification, aiming for greater identity stability at the cost of increased computational complexity. The trackers were run through BoxMOT\footnote{\url{https://github.com/mikel-brostrom/boxmot}} with its default parameters.

\subsection{Datasets}

\begin{figure}[t]
    \centering
    \includegraphics[width=1\columnwidth]{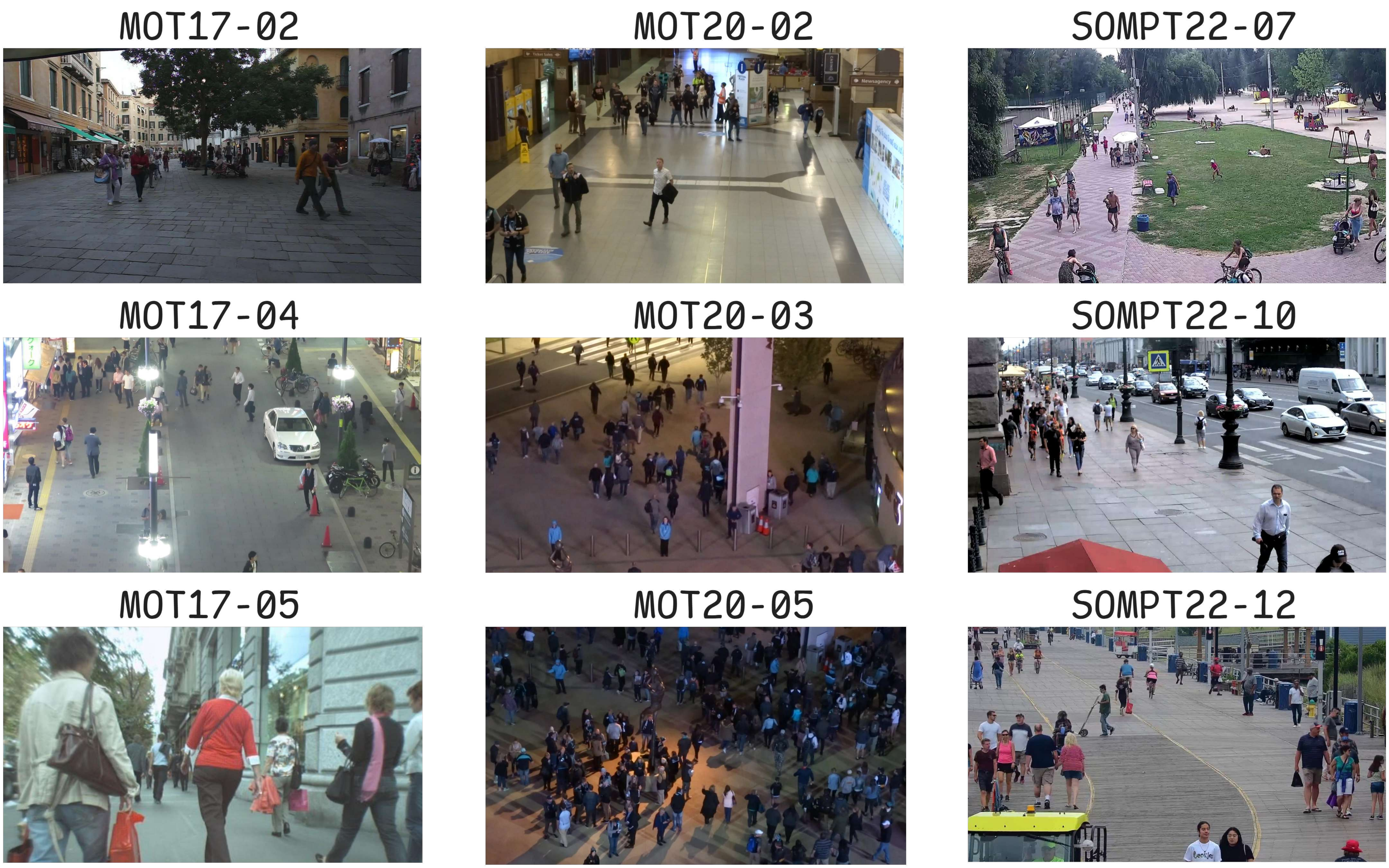}
    \caption{Example frames from the evaluated sequences.}
    \label{fig:frames}
\end{figure}

The datasets were selected to cover complementary pedestrian tracking conditions. MOT17~\cite{dendorfer2020motchallenge} was used as the reference urban benchmark. MOT20~\cite{dendorfer2020mot20} was included because it emphasizes extremely dense scenes with severe occlusions and a large number of people per frame. SOMPT22~\cite{simsek2022sompt22} complements this set with a static surveillance scenario captured by elevated cameras, more closely resembling video surveillance applications. Figure~\ref{fig:frames} illustrates examples of these sequences.

Three sequences were evaluated from each benchmark: 02, 04, and 05 from MOT17; 02, 03, and 05 from MOT20; and 07, 10, and 12 from SOMPT22. The average densities were 28.2 pedestrians per frame for MOT17, 47.7 for SOMPT22, and 149.2 for MOT20. Selecting three sequences per benchmark is consistent with previous studies: Alkandary et al.~\cite{alkandary2025comparative} reported results on three sequences from MOT17 and three from MOT20, and both ByteTrack~\cite{zhang2022bytetrack} and BoT-SORT~\cite{aharon2022botsort} used subsets of MOT17 in their validation experiments.

\subsection{Evaluation Metrics}

The evaluation was conducted using TrackEval\footnote{\url{https://github.com/JonathonLuiten/TrackEval}}. The main metric is HOTA~\cite{luiten2020hota}, which jointly balances detection, temporal association, and localization. As complementary metrics, we report IDF1~\cite{ristani2016performance} to analyze identity preservation and MOTA~\cite{bernardin2008clear} to provide an overview of system errors.

In addition to these aggregate metrics, we use two error counters to aid in the interpretation of results: IDSW (\textit{Identity Switches}), which counts identity switches, and Frag (\textit{Fragmentations}), which indicates trajectory interruptions. For computational efficiency, we report FPS (\textit{Frames Per Second}) and latency.

\subsection{Measurement Protocol}

The experiments were conducted on an NVIDIA RTX 4070 Ti Super GPU, an AMD Ryzen 9 9900X CPU clocked at 4.40 GHz, and 32 GB of DDR5 RAM. FPS and latency values should be interpreted as relative comparisons within this environment, as absolute values may vary with hardware.

Each detector-tracker configuration was run five times on every sequence, across the available scales for each family and, in the case of YOLO26, in both inference modes. Tracking quality metrics (HOTA, IDF1, MOTA, IDSW, Frag) showed only marginal variation across runs (an average variation of approximately 0.03 HOTA points) and are reported as the mean over runs and sequences of each benchmark, which explains the non-integer IDSW and Frag values. FPS and latency are reported as mean and standard deviation across the five runs, with FPS given by 1000/latency.

%=============================================================
\section{RESULTS AND DISCUSSION}
%=============================================================

\subsection{Overview of Tracking Quality}

\begin{table*}[!thb]
\centering
\caption{Results with ByteTrack.}
\label{tab:quality_bytetrack}
\scriptsize
\setlength{\tabcolsep}{3pt}
\begin{tabular}{l *{15}{c}}
\toprule
\multirow{2}{*}{Model} & \multicolumn{5}{c}{MOT17} & \multicolumn{5}{c}{SOMPT22} & \multicolumn{5}{c}{MOT20} \\
\cmidrule(lr){2-6} \cmidrule(lr){7-11} \cmidrule(lr){12-16}
 & HOTA$\uparrow$ & IDF1$\uparrow$ & MOTA$\uparrow$ & IDSW$\downarrow$ & Frag$\downarrow$
 & HOTA$\uparrow$ & IDF1$\uparrow$ & MOTA$\uparrow$ & IDSW$\downarrow$ & Frag$\downarrow$
 & HOTA$\uparrow$ & IDF1$\uparrow$ & MOTA$\uparrow$ & IDSW$\downarrow$ & Frag$\downarrow$ \\
\midrule
YOLO11-N       & 34.43 & 40.20 & 30.24 & 44.80 & 81.27 & 28.10 & 31.76 & 28.46 & 114.47 & 202.20 & 12.60 & 15.72 & 10.21 & 129.80 & 366.07 \\
YOLO11-S       & 36.09 & 42.94 & 32.07 & 48.47 & 95.93 & 30.85 & 35.03 & 31.99 & 113.73 & 229.53 & 12.37 & 14.60 &  9.51 & 131.00 & 331.20 \\
YOLO11-M       & 36.94 & 43.48 & 32.35 & 51.67 & 93.80 & 30.95 & 34.37 & 31.47 &  98.67 & 192.07 & 11.24 & 12.74 &  8.17 &  67.93 & 151.13 \\
YOLO11-L       & 36.13 & 42.88 & 32.29 & 44.40 & 86.93 & 30.24 & 34.09 & 29.43 &  83.67 & 167.00 & 10.07 & 11.14 &  6.73 &  36.40 &  91.67 \\
YOLO11-X       & 36.56 & 43.00 & 31.48 & 43.33 & 83.00 & 31.32 & 35.94 & 30.99 &  81.00 & 171.27 & 10.67 & 11.64 &  6.99 &  48.73 & 100.07 \\
\midrule
RT-DETR-L     & 38.37 & 47.15 & 36.15 & 68.07 & 122.73 & 34.11 & \textbf{39.16} & 36.23 & 100.93 & 201.67 & 16.85 & 20.32 & 13.17 & 130.47 & 282.67 \\
RT-DETR-X     & \textbf{39.20} & \textbf{48.71} & \textbf{37.11} & 70.67 & 116.67 & \textbf{34.18} & 38.88 & \textbf{36.92} &  97.93 & 195.80 & \textbf{17.64} & \textbf{21.96} & \textbf{14.32} & 138.60 & 301.13 \\
\midrule
YOLO26-N (NMS) & 32.63 & 37.53 & 28.25 & 45.00 & \textbf{77.33} & 26.79 & 30.59 & 26.74 & 100.67 & 169.93 &  9.28 & 10.65 &  6.33 &  72.87 & 171.40 \\
YOLO26-N (E2E) & 32.54 & 37.29 & 27.53 & 46.00 & 111.80 & 26.00 & 29.29 & 24.97 &  98.93 & 251.67 &  8.82 &  9.89 &  6.02 &  82.67 & 287.27 \\
YOLO26-S (NMS) & 35.98 & 42.50 & 31.48 & 43.40 & 89.53 & 31.26 & 35.22 & 32.18 & 104.40 & 219.13 &  9.32 & 10.59 &  6.55 &  47.00 & 107.20 \\
YOLO26-S (E2E) & 35.84 & 42.35 & 30.70 & 46.00 & 104.40 & 29.95 & 33.72 & 30.42 & 104.27 & 305.60 &  9.24 & 10.39 &  6.56 &  44.13 & 140.13 \\
YOLO26-M (NMS) & 36.92 & 43.91 & 31.90 & 44.00 & 93.27 & 30.35 & 34.01 & 29.80 &  92.27 & 181.53 &  9.76 & 10.97 &  6.90 &  36.87 & \textbf{84.93} \\
YOLO26-M (E2E) & 36.38 & 43.44 & 31.43 & 42.73 & 110.13 & 28.88 & 32.30 & 27.94 &  97.07 & 240.80 &  8.81 &  9.90 &  6.14 &  \textbf{35.67} & 113.20 \\
YOLO26-L (NMS) & 37.11 & 44.18 & 31.79 & \textbf{42.27} & 90.40 & 31.95 & 36.25 & 31.59 &  98.13 & 192.53 & 11.01 & 12.53 &  7.62 &  45.93 &  97.67 \\
YOLO26-L (E2E) & 36.76 & 43.69 & 31.18 & 46.27 & 142.13 & 30.66 & 34.48 & 29.89 & 106.13 & 370.27 &  9.98 & 11.20 &  6.66 &  52.33 & 283.93 \\
YOLO26-X (NMS) & 37.84 & 45.37 & 32.94 & 48.00 & 91.33 & 30.89 & 34.87 & 28.98 & \textbf{75.20} & \textbf{149.93} & 11.86 & 13.30 &  7.97 &  47.33 &  97.80 \\
YOLO26-X (E2E) & 36.18 & 43.24 & 32.00 & 51.67 & 175.73 & 29.00 & 32.28 & 26.38 &  83.07 & 287.73 & 10.36 & 11.19 &  6.63 &  47.80 & 295.60 \\
\bottomrule
\end{tabular}
\end{table*}

\begin{table*}[!thb]
\centering
\caption{Results with BoT-SORT.}
\label{tab:quality_botsort}
\scriptsize
\setlength{\tabcolsep}{3pt}
\begin{tabular}{l *{15}{c}}
\toprule
\multirow{2}{*}{Model} & \multicolumn{5}{c}{MOT17} & \multicolumn{5}{c}{SOMPT22} & \multicolumn{5}{c}{MOT20} \\
\cmidrule(lr){2-6} \cmidrule(lr){7-11} \cmidrule(lr){12-16}
 & HOTA$\uparrow$ & IDF1$\uparrow$ & MOTA$\uparrow$ & IDSW$\downarrow$ & Frag$\downarrow$
 & HOTA$\uparrow$ & IDF1$\uparrow$ & MOTA$\uparrow$ & IDSW$\downarrow$ & Frag$\downarrow$
 & HOTA$\uparrow$ & IDF1$\uparrow$ & MOTA$\uparrow$ & IDSW$\downarrow$ & Frag$\downarrow$ \\
\midrule
YOLO11-N       & 34.46 & 39.30 & 28.58 & 22.67 &  66.07 & 27.53 & 31.36 & 24.30 & 47.87 & 115.27 & 10.75 & 12.36 &  7.47 & 34.53 & 135.13 \\
YOLO11-S       & 36.63 & 42.35 & 31.11 & 22.33 &  70.13 & 30.34 & 34.85 & 28.14 & 54.00 & 137.33 & 10.88 & 12.11 &  7.76 & 44.67 & 150.20 \\
YOLO11-M       & 36.89 & 42.88 & 31.38 & 24.67 &  65.00 & 30.33 & 34.27 & 27.37 & 46.60 & 117.20 &  9.65 & 10.14 &  6.36 & 22.80 &  67.47 \\
YOLO11-L       & 37.85 & 43.87 & 30.92 & 24.20 &  66.80 & 29.85 & 33.59 & 25.45 & 36.67 & 104.67 &  8.57 &  8.68 &  5.12 & \textbf{9.67} &  40.73 \\
YOLO11-X       & 36.64 & 42.21 & 31.01 & 27.00 &  75.00 & 29.92 & 34.37 & 26.75 & 39.07 & 105.27 &  9.03 &  8.98 &  5.27 & 11.33 &  38.87 \\
\midrule
RT-DETR-L     & \textbf{39.03} & \textbf{46.86} & 34.75 & 36.73 &  90.40 & 31.63 & 36.40 & 27.55 & 35.93 & 114.27 & 11.69 & 12.17 &  7.21 & 21.33 &  76.60 \\
RT-DETR-X     & 38.63 & 46.48 & \textbf{35.55} & 35.47 &  90.80 & \textbf{32.09} & \textbf{36.43} & 27.97 & 36.00 & 104.33 & \textbf{13.16} & \textbf{14.07} & \textbf{8.64} & 29.27 &  98.33 \\
\midrule
YOLO26-N (NMS) & 33.43 & 37.11 & 26.54 & 21.33 & \textbf{57.93} & 25.73 & 29.38 & 22.48 & 44.67 &  93.93 &  7.59 &  7.62 &  4.45 & 17.00 &  60.93 \\
YOLO26-N (E2E) & 32.63 & 36.57 & 26.92 & 27.60 &  90.93 & 25.72 & 28.99 & 22.70 & 52.73 & 168.53 &  7.95 &  8.39 &  5.05 & 26.27 & 136.80 \\
YOLO26-S (NMS) & 37.01 & 43.14 & 30.41 & 27.20 &  76.13 & 31.16 & 35.40 & \textbf{28.26} & 45.53 & 122.40 &  8.12 &  8.63 &  5.26 & 18.00 &  49.40 \\
YOLO26-S (E2E) & 36.58 & 42.70 & 30.77 & 29.87 &  87.33 & 29.88 & 34.02 & 27.30 & 47.60 & 193.67 &  8.18 &  8.56 &  5.31 & 17.73 &  70.80 \\
YOLO26-M (NMS) & 38.08 & 44.50 & 30.98 & 25.67 &  73.33 & 29.77 & 33.59 & 25.87 & 40.93 & 104.27 &  8.03 &  8.58 &  5.29 & 16.33 &  42.13 \\
YOLO26-M (E2E) & 38.13 & 44.90 & 31.53 & 24.00 &  94.00 & 28.71 & 32.33 & 24.74 & 40.67 & 150.27 &  7.54 &  8.10 &  4.99 & 12.80 &  54.80 \\
YOLO26-L (NMS) & 38.19 & 44.55 & 31.12 & \textbf{20.67} &  71.67 & 31.79 & 36.05 & 27.47 & 41.67 & 112.33 &  9.33 &  9.82 &  5.81 & 12.67 &  45.73 \\
YOLO26-L (E2E) & 37.60 & 43.93 & 30.91 & 23.53 & 115.13 & 29.60 & 33.23 & 26.18 & 51.60 & 239.27 &  8.54 &  9.00 &  5.20 & 18.33 & 141.73 \\
YOLO26-X (NMS) & 37.90 & 44.14 & 32.42 & 25.33 &  73.53 & 29.37 & 33.16 & 25.01 & \textbf{34.67} & \textbf{93.27} &  9.63 &  9.48 &  5.61 & 12.00 & \textbf{38.27} \\
YOLO26-X (E2E) & 37.16 & 43.52 & 31.80 & 28.80 & 136.53 & 28.31 & 31.59 & 23.49 & 41.33 & 187.20 &  8.45 &  8.36 &  4.84 & 13.27 & 127.33 \\
\bottomrule
\end{tabular}
\end{table*}

Tables~\ref{tab:quality_bytetrack} and~\ref{tab:quality_botsort} show that the RT-DETR family achieved the highest HOTA on every benchmark with both trackers, especially in the higher-capacity configurations. With ByteTrack, RT-DETR-X achieved the best HOTA across all three benchmarks: 39.20 on MOT17, 34.18 on SOMPT22, and 17.64 on MOT20. With BoT-SORT, RT-DETR-L led in MOT17 (39.03), while RT-DETR-X achieved the highest values in SOMPT22 (32.09) and MOT20 (13.16). Its advantage was clearest in the denser benchmarks: with both trackers, the HOTA margin of RT-DETR-X over YOLO26-X with NMS grew from MOT17 to SOMPT22 and MOT20.

At the same time, stronger overall performance does not necessarily correspond to lower values across all error counters. In several configurations, especially in YOLO26-L and YOLO26-X with NMS, IDSW and Frag were lower than those of RT-DETR despite lower HOTA values. These results show that detector comparisons depend on the metric considered and that error counters should be interpreted together with aggregate quality metrics.

\subsection{Effect of Scene Density and Model Scale}

Across benchmarks, we observe not only a general drop in performance but also a shift in the relative advantage among detector families. In all configurations, HOTA and IDF1 decline from MOT17 to SOMPT22 and, even more markedly, to MOT20. However, this degradation is less severe for RT-DETR. With ByteTrack, the drop in HOTA between MOT17 and MOT20 was 55.0\% for RT-DETR-X, compared to 68.7\% for YOLO26-X with NMS and 70.8\% for YOLO11-X. With BoT-SORT, the pattern repeats: the reduction was 65.9\% for RT-DETR-X, 74.6\% for YOLO26-X with NMS, and 75.4\% for YOLO11-X. Thus, RT-DETR's relative advantage was more evident in the denser benchmarks, especially MOT20, although these benchmarks also differ in factors other than density. Based on the literature~\cite{zhao2024rtdetr}, a plausible hypothesis is that the global attention of RT-DETR’s hybrid encoder models the spatial context between objects and thus reduces ambiguities among pedestrians under severe occlusion. However, this work does not experimentally isolate the detector’s internal mechanisms, so this hypothesis cannot be confirmed causally.

In YOLO11 with ByteTrack, moving from \textit{nano} to \textit{medium} improves HOTA on MOT17, but the gains stop growing at larger scales and are reversed on MOT20. In YOLO26, the highest mean HOTA within the family was achieved by a \textit{large} or \textit{extra-large} variant with NMS in all six benchmark--tracker combinations. Within the end-to-end regime, however, BoT-SORT achieved its highest YOLO26 HOTA with the \textit{medium} variant on MOT17 and the \textit{small} variant on SOMPT22. These results show that increasing model scale does not guarantee higher HOTA. The benefits of larger variants depend on the detector family, benchmark, and tracker used.

Error counters exhibit different patterns across models and benchmarks. Although MOT20 has the lowest HOTA values, Frag does not increase uniformly across all configurations. The reported metrics therefore show the performance decline on this benchmark but do not establish the separate contributions of detection and association to that decline.

\subsection{Comparison of ByteTrack and BoT-SORT}

Replacing ByteTrack with BoT-SORT did not substantially alter the overall hierarchy among the detector families, but it changed the result profile. In all configurations, BoT-SORT yielded lower IDSW and Frag values. On MOT20, for example, replacing ByteTrack with BoT-SORT when using RT-DETR-X reduced IDSW from 138.60 to 29.27 and Frag from 301.13 to 98.33. Nevertheless, these reductions did not consistently translate into higher HOTA. On SOMPT22 and MOT20, ByteTrack achieved higher HOTA in every configuration. Thus, ByteTrack stood out for its speed and aggregate quality on these benchmarks, whereas BoT-SORT yielded lower counts of identity switches and fragmentations.

\begin{table*}[!thb]
\centering
\caption{Computational efficiency.}
\label{tab:efficiency}
\scriptsize
\setlength{\tabcolsep}{2.5pt}
\resizebox{\textwidth}{!}{
\begin{tabular}{lcccccccccccc}
\toprule
\multirow{3}{*}{Model} 
& \multicolumn{4}{c}{MOT17} 
& \multicolumn{4}{c}{SOMPT22} 
& \multicolumn{4}{c}{MOT20} \\
\cmidrule(lr){2-5} \cmidrule(lr){6-9} \cmidrule(lr){10-13}
& \multicolumn{2}{c}{ByteTrack} & \multicolumn{2}{c}{BoT-SORT}
& \multicolumn{2}{c}{ByteTrack} & \multicolumn{2}{c}{BoT-SORT}
& \multicolumn{2}{c}{ByteTrack} & \multicolumn{2}{c}{BoT-SORT} \\
\cmidrule(lr){2-3} \cmidrule(lr){4-5}
\cmidrule(lr){6-7} \cmidrule(lr){8-9}
\cmidrule(lr){10-11} \cmidrule(lr){12-13}
& FPS$\uparrow$ & Lat. (ms)$\downarrow$ & FPS$\uparrow$ & Lat. (ms)$\downarrow$
& FPS$\uparrow$ & Lat. (ms)$\downarrow$ & FPS$\uparrow$ & Lat. (ms)$\downarrow$
& FPS$\uparrow$ & Lat. (ms)$\downarrow$ & FPS$\uparrow$ & Lat. (ms)$\downarrow$ \\
\midrule
YOLO11-N      & 96.1{\scriptsize$\pm$9.1} & 10.4{\scriptsize$\pm$1.1} & \textbf{34.7}{\scriptsize$\pm$2.6} & \textbf{28.8}{\scriptsize$\pm$2.5} & 66.1{\scriptsize$\pm$6.2} & 15.1{\scriptsize$\pm$1.4} & \textbf{27.8}{\scriptsize$\pm$1.6} & \textbf{36.0}{\scriptsize$\pm$1.9} & 80.2{\scriptsize$\pm$7.8} & 12.5{\scriptsize$\pm$1.2} & 30.9{\scriptsize$\pm$2.4} & 32.4{\scriptsize$\pm$2.3} \\
YOLO11-S      & 99.1{\scriptsize$\pm$5.0} & 10.1{\scriptsize$\pm$0.7} & 33.7{\scriptsize$\pm$2.1} & 29.7{\scriptsize$\pm$2.2} & 65.2{\scriptsize$\pm$4.2} & 15.3{\scriptsize$\pm$1.0} & 26.7{\scriptsize$\pm$1.4} & 37.5{\scriptsize$\pm$1.7} & 83.1{\scriptsize$\pm$5.8} & 12.0{\scriptsize$\pm$0.9} & 30.3{\scriptsize$\pm$3.3} & 33.0{\scriptsize$\pm$3.7} \\
YOLO11-M      & 90.4{\scriptsize$\pm$4.0} & 11.1{\scriptsize$\pm$0.6} & 32.0{\scriptsize$\pm$2.2} & 31.2{\scriptsize$\pm$2.5} & 65.6{\scriptsize$\pm$1.8} & 15.2{\scriptsize$\pm$0.4} & 25.4{\scriptsize$\pm$1.6} & 39.3{\scriptsize$\pm$2.1} & 83.4{\scriptsize$\pm$3.3} & 12.0{\scriptsize$\pm$0.5} & 30.5{\scriptsize$\pm$3.1} & 32.8{\scriptsize$\pm$4.0} \\
YOLO11-L      & 77.3{\scriptsize$\pm$2.1} & 12.9{\scriptsize$\pm$0.4} & 30.6{\scriptsize$\pm$2.0} & 32.6{\scriptsize$\pm$2.5} & 57.9{\scriptsize$\pm$2.0} & 17.3{\scriptsize$\pm$0.6} & 24.1{\scriptsize$\pm$1.9} & 41.5{\scriptsize$\pm$3.0} & 69.4{\scriptsize$\pm$4.2} & 14.4{\scriptsize$\pm$0.9} & 30.4{\scriptsize$\pm$2.6} & 32.9{\scriptsize$\pm$3.2} \\
YOLO11-X      & 70.3{\scriptsize$\pm$3.3} & 14.2{\scriptsize$\pm$0.7} & 29.1{\scriptsize$\pm$2.0} & 34.4{\scriptsize$\pm$2.0} & 56.5{\scriptsize$\pm$1.5} & 17.7{\scriptsize$\pm$0.5} & 23.9{\scriptsize$\pm$0.9} & 41.9{\scriptsize$\pm$1.6} & 64.3{\scriptsize$\pm$2.6} & 15.6{\scriptsize$\pm$0.6} & 29.2{\scriptsize$\pm$2.5} & 34.2{\scriptsize$\pm$3.0} \\
\midrule
RT-DETR-L     & 46.2{\scriptsize$\pm$1.1} & 21.7{\scriptsize$\pm$0.6} & 23.9{\scriptsize$\pm$0.9} & 41.9{\scriptsize$\pm$1.7} & 37.2{\scriptsize$\pm$1.3} & 26.9{\scriptsize$\pm$1.0} & 20.8{\scriptsize$\pm$0.9} & 48.1{\scriptsize$\pm$2.2} & 44.0{\scriptsize$\pm$2.1} & 22.7{\scriptsize$\pm$1.1} & 23.6{\scriptsize$\pm$1.7} & 42.3{\scriptsize$\pm$2.8} \\
RT-DETR-X     & 38.0{\scriptsize$\pm$0.6} & 26.3{\scriptsize$\pm$0.5} & 22.1{\scriptsize$\pm$0.8} & 45.3{\scriptsize$\pm$1.7} & 33.5{\scriptsize$\pm$0.9} & 29.8{\scriptsize$\pm$0.8} & 19.8{\scriptsize$\pm$1.3} & 50.5{\scriptsize$\pm$3.3} & 36.5{\scriptsize$\pm$1.2} & 27.4{\scriptsize$\pm$0.9} & 21.3{\scriptsize$\pm$1.8} & 47.0{\scriptsize$\pm$4.1} \\
\midrule
YOLO26-N (NMS) & 99.2{\scriptsize$\pm$10.1} & 10.1{\scriptsize$\pm$0.8} & 33.1{\scriptsize$\pm$1.8} & 30.2{\scriptsize$\pm$1.7} & 67.1{\scriptsize$\pm$4.0} & 14.9{\scriptsize$\pm$0.9} & 27.5{\scriptsize$\pm$1.6} & 36.4{\scriptsize$\pm$2.1} & 87.6{\scriptsize$\pm$3.5} & 11.4{\scriptsize$\pm$0.4} & 33.3{\scriptsize$\pm$3.6} & 30.0{\scriptsize$\pm$3.5} \\
YOLO26-N (E2E) & \textbf{99.9}{\scriptsize$\pm$4.6} & \textbf{10.0}{\scriptsize$\pm$0.5} & 34.3{\scriptsize$\pm$2.0} & 29.1{\scriptsize$\pm$1.9} & \textbf{70.0}{\scriptsize$\pm$3.1} & \textbf{14.3}{\scriptsize$\pm$0.6} & 27.4{\scriptsize$\pm$1.4} & 36.5{\scriptsize$\pm$1.8} & 87.4{\scriptsize$\pm$5.5} & 11.4{\scriptsize$\pm$0.7} & 33.3{\scriptsize$\pm$3.2} & 30.0{\scriptsize$\pm$3.0} \\
YOLO26-S (NMS) & 96.3{\scriptsize$\pm$2.3} & 10.4{\scriptsize$\pm$0.3} & 32.3{\scriptsize$\pm$1.5} & 30.9{\scriptsize$\pm$1.5} & 65.3{\scriptsize$\pm$3.5} & 15.3{\scriptsize$\pm$0.8} & 26.3{\scriptsize$\pm$1.5} & 38.1{\scriptsize$\pm$2.2} & 87.6{\scriptsize$\pm$2.8} & 11.4{\scriptsize$\pm$0.3} & 35.5{\scriptsize$\pm$2.5} & 28.2{\scriptsize$\pm$2.1} \\
YOLO26-S (E2E) & 90.6{\scriptsize$\pm$4.2} & 11.0{\scriptsize$\pm$0.5} & 34.0{\scriptsize$\pm$0.7} & 29.4{\scriptsize$\pm$0.6} & 66.4{\scriptsize$\pm$2.9} & 15.1{\scriptsize$\pm$0.7} & 26.4{\scriptsize$\pm$1.5} & 37.8{\scriptsize$\pm$2.2} & \textbf{91.7}{\scriptsize$\pm$3.8} & \textbf{10.9}{\scriptsize$\pm$0.4} & \textbf{35.6}{\scriptsize$\pm$2.0} & \textbf{28.1}{\scriptsize$\pm$1.8} \\
YOLO26-M (NMS) & 89.1{\scriptsize$\pm$4.6} & 11.2{\scriptsize$\pm$0.5} & 30.9{\scriptsize$\pm$1.6} & 32.4{\scriptsize$\pm$2.0} & 66.0{\scriptsize$\pm$1.2} & 15.2{\scriptsize$\pm$0.3} & 25.5{\scriptsize$\pm$1.4} & 39.2{\scriptsize$\pm$2.1} & 85.3{\scriptsize$\pm$2.7} & 11.7{\scriptsize$\pm$0.3} & 34.0{\scriptsize$\pm$2.4} & 29.4{\scriptsize$\pm$2.7} \\
YOLO26-M (E2E) & 89.0{\scriptsize$\pm$8.9} & 11.2{\scriptsize$\pm$1.0} & 31.2{\scriptsize$\pm$2.4} & 32.1{\scriptsize$\pm$3.0} & 67.5{\scriptsize$\pm$1.2} & 14.8{\scriptsize$\pm$0.3} & 25.5{\scriptsize$\pm$1.9} & 39.2{\scriptsize$\pm$2.8} & 86.6{\scriptsize$\pm$3.1} & 11.5{\scriptsize$\pm$0.4} & 35.4{\scriptsize$\pm$1.8} & 28.3{\scriptsize$\pm$2.3} \\
YOLO26-L (NMS) & 74.8{\scriptsize$\pm$2.6} & 13.4{\scriptsize$\pm$0.5} & 30.5{\scriptsize$\pm$0.4} & 32.8{\scriptsize$\pm$0.4} & 58.7{\scriptsize$\pm$2.0} & 17.0{\scriptsize$\pm$0.6} & 25.7{\scriptsize$\pm$1.1} & 38.9{\scriptsize$\pm$1.6} & 72.1{\scriptsize$\pm$2.7} & 13.9{\scriptsize$\pm$0.5} & 29.7{\scriptsize$\pm$2.9} & 33.7{\scriptsize$\pm$3.7} \\
YOLO26-L (E2E) & 73.8{\scriptsize$\pm$3.9} & 13.6{\scriptsize$\pm$0.9} & 30.4{\scriptsize$\pm$0.5} & 32.9{\scriptsize$\pm$0.6} & 57.3{\scriptsize$\pm$3.3} & 17.5{\scriptsize$\pm$1.0} & 25.6{\scriptsize$\pm$1.2} & 39.1{\scriptsize$\pm$1.8} & 70.4{\scriptsize$\pm$1.3} & 14.2{\scriptsize$\pm$0.3} & 30.2{\scriptsize$\pm$2.8} & 33.1{\scriptsize$\pm$3.4} \\
YOLO26-X (NMS) & 71.9{\scriptsize$\pm$2.8} & 13.9{\scriptsize$\pm$0.6} & 29.2{\scriptsize$\pm$1.2} & 34.3{\scriptsize$\pm$1.6} & 58.5{\scriptsize$\pm$1.5} & 17.1{\scriptsize$\pm$0.5} & 25.1{\scriptsize$\pm$0.4} & 39.9{\scriptsize$\pm$0.6} & 64.3{\scriptsize$\pm$3.0} & 15.6{\scriptsize$\pm$0.8} & 29.5{\scriptsize$\pm$1.9} & 33.9{\scriptsize$\pm$1.9} \\
YOLO26-X (E2E) & 74.4{\scriptsize$\pm$2.1} & 13.4{\scriptsize$\pm$0.4} & 28.9{\scriptsize$\pm$1.1} & 34.6{\scriptsize$\pm$1.4} & 59.6{\scriptsize$\pm$1.1} & 16.8{\scriptsize$\pm$0.3} & 24.8{\scriptsize$\pm$0.8} & 40.3{\scriptsize$\pm$1.3} & 67.9{\scriptsize$\pm$1.6} & 14.7{\scriptsize$\pm$0.3} & 30.9{\scriptsize$\pm$1.4} & 32.4{\scriptsize$\pm$1.3} \\
\bottomrule
\end{tabular}
}
\end{table*}

\subsection{Trade-off between Quality and Computational Efficiency}

Table~\ref{tab:efficiency} shows that computational efficiency varies systematically depending on the detector, the scale, and the tracker. ByteTrack was consistently faster than BoT-SORT across all configurations, in line with the appearance information and motion compensation that BoT-SORT adds.

Among the detectors, RT-DETR had the lowest FPS values, while the highest values were achieved by the \textit{nano} and \textit{small} variants of YOLO11 and YOLO26, depending on the benchmark and tracker. In MOT17 with ByteTrack, for example, RT-DETR-X achieved 38.0 FPS, whereas YOLO26-X with NMS achieved 71.9 FPS and YOLO26-M with NMS reached 89.1 FPS. This comparison illustrates the trade-off between HOTA and speed in the evaluated configurations. However, FPS rankings do not directly follow FLOP counts: RT-DETR-L, for example, achieved lower FPS than the \textit{extra-large} variants of YOLO11 and YOLO26 despite having fewer FLOPs (Table~\ref{tab:models}).

The \textit{nano} and \textit{small} variants are also candidates when processing speed is a priority. On MOT17 with ByteTrack, YOLO11-N, YOLO11-S, YOLO26-N (NMS), and YOLO26-S (NMS) achieved approximately 100 FPS, well above RT-DETR, but with a clear loss in HOTA. Since the experiments were run with public weights without fine-tuning, additional training targeted at the smaller models may narrow this gap.

On MOT17 and SOMPT22, different YOLO26 variants offered favorable combinations of HOTA and speed compared with RT-DETR. The most advantageous scale varied, but across these two benchmarks and both trackers, YOLO26-L with NMS was 0.9--6.5\% below the RT-DETR variant with the highest mean HOTA in each benchmark--tracker combination, while achieving 27.6--96.8\% higher FPS. On MOT17 with ByteTrack, YOLO26-X with NMS was 3.5\% below RT-DETR-X in HOTA (37.84 vs. 39.20), but was 89.2\% faster in FPS (71.9{$\pm$2.8} vs. 38.0{$\pm$0.6}). On MOT17 with BoT-SORT, YOLO26-M with NMS was 2.4\% below RT-DETR-L (38.08 vs. 39.03) and was 29.3\% faster (30.9{$\pm$1.6} vs. 23.9{$\pm$0.9}). On SOMPT22 with BoT-SORT, YOLO26-L with NMS was only 0.9\% below RT-DETR-X (31.79 vs. 32.09), with a 29.8\% gain in FPS (25.7{$\pm$1.1} vs. 19.8{$\pm$1.3}). In practical terms, the results highlight smaller variants as candidates when speed is a priority, different YOLO26 configurations as alternatives for balancing HOTA and speed, and RT-DETR when the priority is to maximize HOTA in the evaluated scenarios.

\subsection{NMS and End-to-End Ablation in YOLO26}

The YOLO26 ablation evaluates two inference regimes within the same detector family. With ByteTrack, the NMS-based regime achieved higher mean HOTA than the end-to-end mode on all three benchmarks at every evaluated YOLO26 scale. Averaged across the five evaluated scales, the HOTA advantage of the NMS-based regime was 0.56 points on MOT17, 1.35 on SOMPT22, and 0.80 on MOT20. In the same comparison, the end-to-end mode resulted in mean increases of 40.5, 108.6, and 112.2 fragmentations, respectively.

With BoT-SORT, the trend persisted. Although there are isolated exceptions, the end-to-end mode did not yield consistent tracking gains and again increased Frag at all scales. FPS differences between the two regimes were small and did not consistently favor the end-to-end mode. In contrast, this mode showed recurring HOTA losses and increased fragmentation.

In summary, the NMS-based regime generally achieved higher HOTA and lower fragmentation than the end-to-end mode in the YOLO26 experiments. Speed differences were small and did not consistently favor either regime. The results therefore do not indicate a systematic advantage of end-to-end inference for tracking under the evaluated conditions. Possible causes, such as lower temporal stability or a different confidence distribution of the end-to-end outputs, were not investigated here.

%=============================================================
\section{CONCLUSION}
%=============================================================

This study analyzed, within a single experimental protocol, how detector architecture, model scale, scene density, and inference regime affect pedestrian tracking. The results showed that the RT-DETR family achieved the highest HOTA on every benchmark with both trackers, with a clearer advantage in the more challenging benchmarks, especially in MOT20.

In the tracker comparison, ByteTrack was consistently faster and achieved higher HOTA on SOMPT22 and MOT20 in every configuration, while BoT-SORT yielded lower counts of identity switches and fragmentations. In the YOLO26 ablation, the NMS-based regime generally achieved higher HOTA and lower fragmentation than end-to-end inference. Speed gains in end-to-end mode were small and inconsistent.

Within the evaluated protocol, RT-DETR is preferable when HOTA is the priority, particularly on MOT20. YOLO26-L with NMS offers a favorable balance between HOTA and speed on MOT17 and SOMPT22, while the \textit{nano} and \textit{small} variants remain candidates when speed dominates.

Future work may extend the evaluation to more sequences and benchmarks, such as DanceTrack~\cite{sun2022dancetrack}. It may also include two-stage detectors such as Faster R-CNN~\cite{ren2017fasterrcnn}, fine-tuned detectors, and additional trackers. It may also repeat the NMS and end-to-end comparison on CPU to assess the relative latency of the two inference regimes in that environment.

\section*{Acknowledgements}
The authors thank \textit{Fundação Coordenação de Aperfeiçoamento de Pessoal de Nível Superior (Capes)} for partial financial support for this work -- Finance Code 001.

\end{document}